# Optimal service resource management strategy for IoT-based health information system considering value co-creation of users

Ji Fang[1,2], Vincent CS Lee[2], and Haiyan Wang[1]

[1]School of Economics and Management, Southeast University, Nanjing, China

[2]Department of Data Science and Artificial Intelligence, Monash University, Melbourne, Australia

**Corresponding author:**

Haiyan Wang, School of Economics and Management, Southeast University, Nanjing, 211189, China. Email: hywang@seu.edu.cn

**Abstract**

**Purpose** - This paper explores optimal service resource management strategy, a continuous challenge for health information service to enhance service performance, optimize service resource utilization and deliver interactive health information service.

**Design/methodology/approach** - An adaptive optimal service resource management strategy was developed considering value co-creation model in health information service with a focus on collaborative and interactive with users. The deep reinforcement learning algorithm was embedded in the IoT-based health information service system (I-HISS) to allocate service resource by controlling service provision and service adaptation based on user engagement behaviour. The simulation experiments were conducted to evaluate the significance of the proposed algorithm under different user reactions to the health information service.

**Findings** - The results indicate that the proposed service resource management strategy considering the user co-creation in the service delivery process improved both the service provider's business revenue and users' individual benefits.

**Practical implications** - The findings may facilitate the design and implementation of health information service that can achieve high user service experience with low service operation costs.

**Originality/value** - This study is among the first to propose service resource management model in I-HISS considering value co-creation of user in the service-dominant logic. The novel artificial intelligence algorithm is developed using deep reinforcement learning method to learn the adaptive service resource management strategy. The results emphasize user engagement in health information service process.

**Paper type** Research paper

# 1. Introduction

The last three years has seen healthcare industry undergoing signification transformations in both business operations and human consciousness due to the COVID-19 pandemic (Choi, 2021) requiring social distance. In particular, the implementation of lockdown policies has compelled fitness centres to close, leading to a shift towards online-based personalized fitness platforms, as users are unable to attend fitness centres. Moreover, the pandemic has increased individuals' awareness of their health, leading to a surge in the sale and usage of digital healthcare devices. This trend has facilitated individuals' more proactive participation in data collection practices, promoting the development of innovative health information services. The increasing popularity of online health information services can be attributed to their convenience and effectiveness. For instance, the well-known online health information service provider "FitTime" (http://www.rjFitTime.com/) began to provide the "Pocket Fat Loss Camp" service using the WeChat Mini Program Platform in 2017. This service includes dietary plans, fitness techniques, and online fitness courses. The service also has online specialized health managers who provide ongoing supervision, analyse users' health performance data, and tailor the personalized health management plans to assist users in achieving scientifically weight loss and fitness management. Users have access to every requirement information for their fitness plans on their smartphones, which gets rid of the need to go to traditional gyms.

Our research focuses on the health information service, an online platform that provides real-time situation-triggered reminders, pushes, and notifications with expert knowledge to assist users in improving their self-health management skills. The target users are those who have enrolled in specific health-related services, such as online fitness courses and meal planning subscriptions. The health information service is tasked with the challenge of ensuring the consistency and accuracy of the fitness data uploaded by users. To solve this obstacle, there is increasing interest in using of Internet of Thing (IoT) structures to enhance user engagement experiences in innovative health information services via automated real-time data collection and instant data transmission (Mishra *et al.*, 2016). One example is the appearance of smart hardware products by the global sports technology company "Beijing Calorie Technology Co., Ltd." (also known as "Keep", https://keep.com/) in 2018. Users can connect the smart hardware to the Keep App, which allows them to access fitness course notifications, track various health-related data, and receive intelligent exercise recommendations and plans. The system analyses users' exercise data to provide more guidance for their training in collaboration with smart wearable devices.

The IoT-based health information service system (I-HISS) connects multiple data sources via the IoT structures increasing the scale and accuracy of data collection. The IoT structures in the health information service encompass a network for the connecting and data exchange between diverse devices These devices include wearables, smartphones, and other appliances that collect health-related data concerning the person. This ecosystem of data collection from diverse physical devices can streamline performance measurement and provide peer-to-peer service content for personalized healthcare management. The performance of the health information service is generated when the users make efforts to improve their health or physical fitness using the service. When assessing the performance of I-HISS, we choose the users' improvements to their health or physical fitness during the service period. The effectiveness of providing high-quality health information service results in increased user satisfaction, which in turn leads to a greater reliance on the service and the generation of favourable referrals. This ultimately boosts profits and helps to retain existing users.

The performance of I-HISS is impacted by two factors. The first aspect is the service quality level. For I-HISS, the service quality level is mainly determined by the experience professional level of fitness coaches and the service technology. For example, some users provided feedback on the course settings in "Keep", saying that the courses were not suitable for their specific situations. To tackle this issue, "Keep" hiring additional full-time trainers and integrate artificial intelligence (AI) into the "smart training plans." This led to multiple improvements, including greater adaptability adjustments and course content replacement, which ultimately improved the current level of service quality. However, higher levels of service quality can lead to a significant increase in service costs. From 2020 to 2022, "Keep" experienced a total gross loss of around 1.6 billion (USD).[1] Therefore, it is essential to explore optimizing service resource management for the sustainable operation of I-HISS.

User engagement is another important aspect that influences the performance of the services provided by I-HISS. In the fitness services business, co-creating value is thought to involve active user participation. Through using "Keep" as an example, the average monthly exercise frequency of active users increased regularly by 4.8 times, 4.3 times, 5.0 times, and 4.1 times, respectively, during the period from 2019 to 2022.[1] This reflects an increased level of user engagement, which correlates with an increased user retention rate for the service. One issue with health information service management is that low resource utilization occurs when user abandons service resource even if they have paid for them during the service (Dunbrack, 2014). For example, Laing et al., (2014) conducted a randomized controlled trial using the MyFitnessPal smartphone app for weight loss in overweight people. However, during the sixth month of the trial, only a small number of users continued to use the app regularly, indicating that not all individuals benefit from health information service. The study suggested that only those who are motivated to track calories and lose weight are likely to benefit from such service. In light of this, it is essential to provide personalized health information services to users only when they are willing to accept them.

Our research focuses on addressing the following research questions, taking into account current practices in health information services and user's changing fitness preference.

(1) How does I-HISS analyse users' fitness data to predict their health information service needs, such as fitness supervision feedback and personalized training courses?

(2) How I-HISS quantifies the service experience for different users during the service, taking into account the types of physical characteristics among the users?

Footnote: 1. https://www1.hkexnews.hk/listedco/listconews/sehk/2023/0630/2023063000204_c.pdf

(3) How can service resources be flexibly allocated at different times in response to fluctuating users' service needs considering the transience characteristic of health information service.

(4) How does I-HISS allocate service resources to optimize the health improvement of users and enhance their overall service experience?

In order to answer these research questions, we deploy the health information service which leverages IoT structure to support service objectives. The IoT structure collects and transmits data from multiple physical devices to the service platform and its users, enabling quantitative analysis of service performance within the service loop. It can improve the service management by monitoring user engagement levels and facilitating interaction between I-HISS and its users. This service mode is popular with online health information service but has seldom been discussed in related research. Our research contributes to existing literature in the following ways.

(1) We adopt a deep reinforcement learning algorithm that combining data-driven and model-driven approaches to predict user engagement behaviour. The adaptability of the algorithm enables it to effectively respond to changes in different environment and the user's health condition.

(2) We integrate the concept of Vargo's value co-creation theory, taking into account the level of user participation in value co-creation and discussing its impact on optimal resource allocation strategies.

(3) We provide management suggestions to improve the service resource allocation strategy of I-HISS, aiming to enhance the effectiveness of health information service.

The rest of the paper is organized as follows. Section 2 covers background and related research. The service resource management problem in IoT-based health information service system (I-HISS) is proposed in Section 3. Section 4 describes the deep reinforcement learning model we proposed to solve the resource management problem in I-HISS. The simulation experiments are presented in Section 5, while the conclusion and future work are given in Section 6.

## 2. Literature review

Service resource management in an IoT-based healthcare information system is controlling service quality effectively to improve the service performance in healthcare services. Understanding the significance of value co-creation is important for optimizing resource management, as collaboration efforts between the service provider and the user to improve the user's health condition. Hence, our study focuses on the service quality of IoT-based health information service, and the value co-creation in the service process. In addition, we review the deep reinforcement learning methods in health information service.

*2.1 Service resource management in IoT-based health information service system*

We focus on three aspects: 1) the development of IoT-based health information service system, 2) the recent research of service resource management in health information service, 3) the challenges of service resource management in health information service.

The health information services provide health-related service through mobile communication devices that enable user interactivity (Liu and Avello, 2020). By expediting access to health information, IoT-based health information service has the potential to revolutionize the delivery of health care. Uses of health information services have increased the availability of health services at reduced total costs (Kao *et al.*, 2018). The practical use of health information service includes health-related physical fitness (Yerrakalva *et al.*, 2019), promoting healthy food (Chen *et al.*, 2018; Grundy *et al.*, 2016).

Previous research in the field of health information services has established a connection between user input and evaluation of service performance. Kranz *et al.*, (2013) highlighted the potential for interactive services to increase acceptability and has the potential to encourage regular exercise among users through personalized sensor-based feedback. With the advent of the Internet of Things (IoT), health information service can provide consumers with highly accurate sensor data for proactive health monitoring (Canhoto and Arp, 2017; Yang *et al.*, 2022). Integrating sensors to a health information service system with sensor network enables the system collection and sharing of data (Fouad *et al.*, 2020; Lomotey *et al.*, 2017). These sensors monitor and record the users' workout data over time, and provide personalized feedback depending on sensor data such as heart rate, location, and acceleration (Stavropoulos *et al.*, 2020). In addition to encouraging self-monitoring, health information services inspire consumers to retain good habits through behavioural contracts, incentives, social support, and contingent benefits (Mohr *et al.*, 2014).

In the research of service resource management, the assessment of service quality is common identified by the mismatch between user expectations and their overall assessment of the service experience (Parasuraman *et al.*, 1988). Existing research emphasize the importance of service quality in maintaining user adoption intentions (Deng *et al.*, 2015). Several models have been developed in the literature to describe and quantify service quality in health service (Chang and Chelladurai, 2003; Lam *et al.*, 2005). It is noteworthy that there is empirical research on consumer involvement in co-creating experiential service quality (Schembri and Sandberg, 2011).

Service resource management in health information services faces multiple challenges, one of which is inadequate research in the health information service market. Prior research has primarily focused on providing descriptions without conducting a comprehensive assessment of the deviations between service resource providing and consumer demands. To address this challenge, we analyse the users' fitness data during the service period to understand the interaction between service providers and users. In addition, there is a lack of evaluation considering the impact of service quality on user service outcomes, despite previous studies highlighting the significance of service quality in healthcare services.

*2.2 Value co-creation in service process*

The extensive related works demonstrates that value co-creation relates to the service process through which users and service providers create value through a collaborative effort. In the service science, users and service providers co-create value by actively engaging in the service process (Lusch and Vargo, 2006; Vargo and Lusch, 2004). In the health information service, the service value is the overall benefit that is derived by the service provider and the users. The service provider and users collaborate to produce value that neither could have achieved alone (Vargo and Lusch, 2008). For the service provider, the value is increased efficiency, cost savings, and revenue growth, which can be measured by metrics like customer retention (Alkitbi *et al.*, 2021; Dadfar *et al.*, 2013; Lin *et al.*, 2009; Oppong *et al.*, 2021; Viglia *et al.*, 2022). For the users, value is for its health and other personal benefits, which can be measured by tracking health or fitness goals or the service's impact on quality of life (Beratarrechea *et al.*, 2014; Medrano-Ureña *et al.*, 2020).

As the level of the user involvement increases, users have a greater impact on the service delivery process and its outcome. This increase in control by the user can have a positive or negative effect, depending on their expertise in co-creation and their perceived value of it (Chen *et al.*, 2017; Shulga and Busser, 2021; Zhang *et al.*, 2020). When the user involves in the service process to create value, users may provide ideas, feedback, and suggestions. The service provider will use their resources and expertise to create the service that meets the user's health needs. After the service interaction, customers are able to evaluate the service and provide feedback to enhance the service in the next period (Agrawal and Rahman, 2015; Saha *et al.*, 2020; Stegmann *et al.*, 2021).

Recent literature highlights the importance of user engagement and value co-creation. However, it also exists two main limitations. Previous studies have frequently mentioned user engagement and value co-creation, but many have not provided quantitative measurement methods. These studies lack defined indicators and standards to assess the level of user engagement and co-creation in service resource management. To fill this gap, this study employs theories from Demirezen *et al.*, (2020), using a modelling approach to quantify users' effort during the value co-creation process. Furthermore, existing literature has simply discussed the importance of value co-creation between users and service providers. However, there has been a lack of detailed research on the specific strategies of this interactive relationship. We fill this gap by exploring the impact of service management strategy through service resource allocation in the value co-creation.

*2.3 Deep reinforcement learning approach in health information service*

Recent research is focused on the applying deep reinforcement learning in health information services. Health information services are significant for changing user behaviour and improving health outcomes. In order to improve the effect of health interventions and minimize the operation costs for the service provider, it is necessary to use artificial intelligence technologies (del Carmen Rodríguez-Hernández and Ilarri, 2021; Gasparetti *et al.*, 2020). By continuously monitoring user behaviours and adjusting service strategies, deep reinforcement learning has provided more efficient and effective management strategies, bringing new opportunities for health information services. We investigate the reality application of deep reinforcement learning in health information services and its potential influence on user experience and health results.

Deep reinforcement learning is an important technique that can monitor user behaviours and tailor the health intervention service depending on user feedback during the service process. This technology allows health information services to dynamically respond to the user's health requirements and changing behaviour, therefore improving the effectiveness of health service. For example, Yom-Tov *et al.* (2017) experimented the deep reinforcement learning approach on 27 sedentary type 2 diabetes patients to encourage them to increase the level of their physical activity. The results showed that the deep reinforcement learning improved gradually in predicting which intervention methods would lead patients to exercise. This experimental research shows the applicability of the deep reinforcement learning in practical situations. Forman *et al.* (2019) evaluated the feasibility, acceptability, cost savings, and effectiveness of the deep reinforcement learning for weight loss intervention in behavioural weight-loss trials. They observed that the reinforcement learning system can considerably lower costs in the optimized conditions while still producing a satisfying intervention effect. This study provides experimental evidence of the practical effect of the deep reinforcement learning in health information management, highlighting its significance in improving system efficiency and user satisfaction.

There is a lack of insight regarding the optimal service management using deep reinforcement learning in health information services. The existing research is inadequate in terms of providing specific and quantitative information in this research area. We aim to address this gap by identifying and quantifying the factors that impact service management. We provide an effective framework for solving the service resource management problem in health information services by using the deep reinforcement learning method. Our objective is to provide insights into how this new technology improves users experience and enhance their health results.

*2.4 Summary of research gaps*

The literature research reveals major research gaps in the aspects of service resource management, value co-creation, and deep reinforcement learning in IoT-based health information services. The gaps include: 1) absence of evaluation indicators to measure user involvement and value co-creation in IoT-based health information service, 2) insufficient understanding of service resource management strategies to promote interactive effect during value co-creation process, and 3) the requirement for definition and quantification in the implementation of deep reinforcement learning in service resource management problem. The aim of our study is to address these gaps by developing a modelling approach that quantifies users' efforts in value co-creation, analysing the impact of service management strategies on service performance during value co-creation process, and providing insights into the practical implementation of deep reinforcement learning in service resource management problem.

## 3. Service resource management problem in IoT-based information service system

In this section, we first provide a brief introduction of the IoT-based health information service system (I-HISS). Thereafter, we clarify the fundamental concepts and describe the service resource management problem in I-HISS.

## 3.1 Overview of the IoT-based information service system

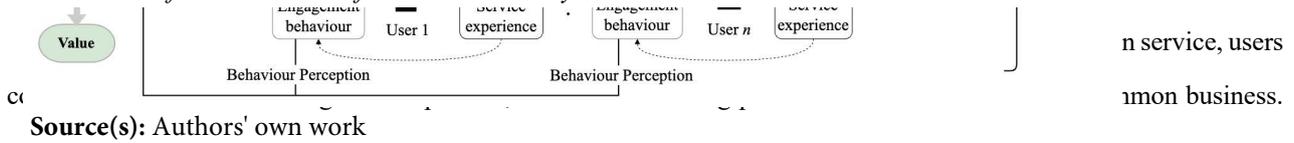
**Source(s):** Authors' own work

Figure *I* depicts the health information service process following the data value chain. Based on the data value chain, the activities in health information service are categorized into four stages. In the first stage, physical devices perceive data sources and transmit them to the service platform, which transforms raw data into information ingredients. In the second stage, processed data are converted into information employing computational resource and expert resource for data analysis tasks. Service platform then provides users with personalized health information in the third stage. Finally, the service value is generated when users utilize the obtained information to improve their health. The service platform is constructed with a three-tier architecture comprising a data tier, a logic tier, and a presentation tier, as proposed by Buschmann (1996). The logic tier in the service platform includes the business logic layer and the service logic layer, which are the core business processes in the I-HISS. The intelligent decision making model is embedded in the business logic layer to support service resource management and provide personalized health services. Due to space limitations, we were unable to include all comprehensive information regarding the context and technologies of the IoT-based health information service platform in this paper. However, we plan to present these findings in future work.

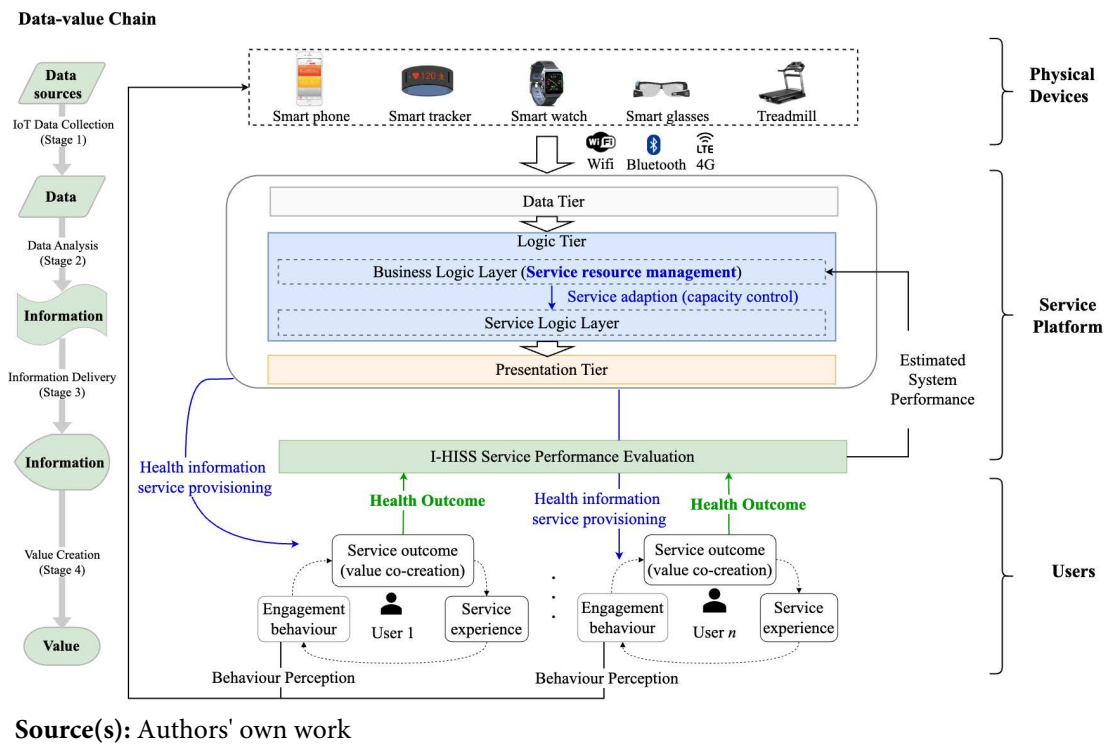
**Source(s):** Authors' own work

Figure I. Overview of the IoT-based health information service system following the data-value chain (left)

The servitization loop within the I-HISS is depicted in

Source(s): Authors' own work

Figure *I*. Users subscribe to the health information service with the aim of achieving specific fitness goals, such as building muscle, losing weight, or maintaining a healthy lifestyle. The service platform provides users with personalized health guidance, such as online fitness courses and healthy eating recommendations, to meet their health needs. The service goal of the I-HISS is to assist users to achieve better health. At each interaction during the service process, the I-HISS collects data from physical devices connected to the user and provides health information service. The service platform makes the resource management of health information service, whereas the service outcome of the I-HISS is co-created by the service platform and the users, particularly when the users it serves exert effort and engage in health-related behaviours (Lim *et al.*, 2018).

*3.2 Service resource management problem in the IoT-based information service system*

This subsection defines essential concepts in the IoT-based health information service system (I-HISS), including control variables related to service resource management problem in the I-HISS and the health outcome as the service objective.

*3.2.1 Control variables in service resource management problem.* This research focuses on two key control variables in resource management: resource provisioning and resource adaptation. Resource provisioning involves that the allocation of a service provider's resource to a user, whereas resource adaptation refers to the capacity of that system to adjust the resource dynamically to fulfill the requirements of the user (Manvi and Krishna Shyam, 2014).

DEFINITION 1. **Service provisioning.** The I-HISS controls the users $i, (i = 1, ..., n)$ who use the health information service. The mathematical formula is:

$$\boldsymbol{y}_t = (y_t^1, y_t^2, ..., y_t^n), \text{ where } y_t^i = \begin{cases} 0, & \text{Not provide the service;} \\ 1, & \text{Provide the service.} \end{cases}$$

The number of users $n_t$ connected to the I-HISS at time $t$ can be calculated as $n_t = \sum_{i=1}^{n} y_t^i$.

DEFINITION 2. **Service adaptation.** The I-HISS dynamically adjust its service capacity to meet the requirements of the users. The available service capacity $c_t$ is a discrete and finite set controlled by the I-HISS at time $t$. The set of alternative service capacity plans is denoted by $c_t \in C = \{C_1, C_2, ..., C_m\}$, where $m$ represents the number of alternative service capacity plans determined by the computing capability of the servers, network bandwidth, healthcare specialists, and other factors.

DEFINITION 3. **Perceived service quality.** The perceived service quality is dependent on both the available service capacity and the number of users who are simultaneously accessing the service. The service quality is assumed to be allocated equally among the accessed users for simplicity of analysis. The mathematical formula of service quality is:

$$q_t = \frac{c_t}{n_t} = \frac{c_t}{\sum_{i=1}^{n} y_t^i} \tag{1}$$

*3.2.2 Objective.* The I-HISS aims to provide health information service to multiple users while ensuring the service resource is allocated effectively. The problem of service resource management problem involves balancing the service quality and resource utilization. On the one hand, optimal service quality perceived by users necessitates the allocation of sufficient resource to users, while on the other, resource is limited, and the utilization aims to reduce the quantity of resource consumed in order to maximize cost efficiency. As a result, the challenge of service resource management is to find the optimal balance between service quality and resource utilization in order to maximize overall performance and user satisfaction.

In this paper, the value of health outcome of all users is considered as the overall service performance, as this reflects the trustworthiness and satisfaction of the users' experience with the health information service. As discussed earlier, the increase in health outcome is because of the collaborative work between the service provider and the user. Therefore, we consider the health outcome as on the Cobb-Douglas production function, which was first introduced by Cobb and Douglas (1928) to model the relationship between production output and production inputs. Similar to the past studies, we consider that the health outcome is a durable capital stock in line with the theory in Grossman (1972). In this theory, health is regarded as an investment with long-term benefits, rather than consumption good. On account of this, health outcomes are viewed valuable not only at the end, but throughout the service period.

DEFINITION 4. **Health outcome.** *The health outcome is defined as the transition function of the health outcome $H_t^i$ for the user $i$ at the decision point $t$, is given by:*

$$H_{t+1}^i = H_t^i + y_t^i (x_t^i)^{\alpha_1} q_t^{\alpha_2} \qquad (2)$$

*Where $\alpha_1, \alpha_2 \in (0,1), \alpha_1 + \alpha_2 < 1$. The parameters $\alpha_1, \alpha_2$ are the service-specific parameters defined by the relative relevance of the user's effort and the IHISS's effort in value co-creation. $\alpha_1 + \alpha_2 < 1$ implies a decreasing return to scale, ensuring that the output of the value co-creation service cannot be infinite.*

In service resource management, allocating service resource to users who exert greater effort can improve service performance. This emphasizes the significance of service provisioning strategy in the optimization of service resource. In addition, the service adaptation strategy attempts to make sure that users obtain the necessary amount of resource without wasting any resource. Therefore, the I-HISS optimizes the resource adaptation trajectory $c_t$ over the service horizon and determines service provisioning of which user pushing health information service $\mathbf{y}_t = (y_t^1, y_t^2, \ldots, y_t^n)$ for all $t$ ($1 \leq t \leq T$). By considering health outcomes as a durable capital stock, the study emphasizes the long-term benefits of health information service and highlights the importance of effective service resource management in maximizing the long-term health benefits. The objective function of the service resource management problem is the accumulation of the user's health outcome at the end of service $T$ for all the users $i, i = 1,2,\ldots,n$, where:

$$\max_{c_t, y_t} \sum_{t=1}^{T} \sum_{i=1}^{n} H_t^i$$
$$s.t. \sum_{t=1}^{T} \beta c_t \leq B. \tag{3}$$

In order to ensure the sustainability of the health information service operations, the I-HISS under a financial constraint represented by the total budget for service operation costs during the service period, denoted as $B$. This financial constraint requires the I-HISS to ensure the service adaptation within the budget and consider the efficacy of service provisioning. $\beta$ represents the cost elasticity of the service capacity. Due to the Cobb-Douglas production function is nonlinear, leading to a nonlinear objective function. In the following section, a deep reinforcement learning approach is developed to solve the nonlinear optimization model.

## 4. Deep reinforcement learning model of service resource management problem

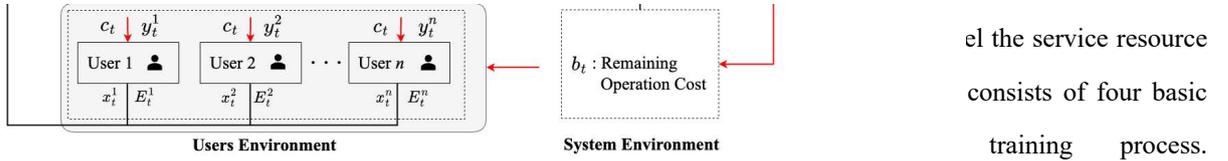

el the service resource consists of four basic training process.

**Source(s):** Authors' own work

Figure *II* depicts the agent-environment interaction in the deep reinforcement learning. At each interaction, the agent chooses service resource management plan $a_t$ from a set of possible actions. $a_t = [c_t, y_t]$ contains service adaptation control $c_t$, the service provisioning decision $y_t$. Therefore, the user perceived service quality $q_t$ can be evaluated using Eq. (1). Each action $a_t$ results in a change in the remaining operation cost in the system environment, as well as the health outcome $H_t^i$ while the users exert efforts $x_t = (x_t^1, x_t^2, \ldots, x_t^n)$ to follow the health guidance and experience changes in their health states that are observable through the physical devices in the consumers environment.

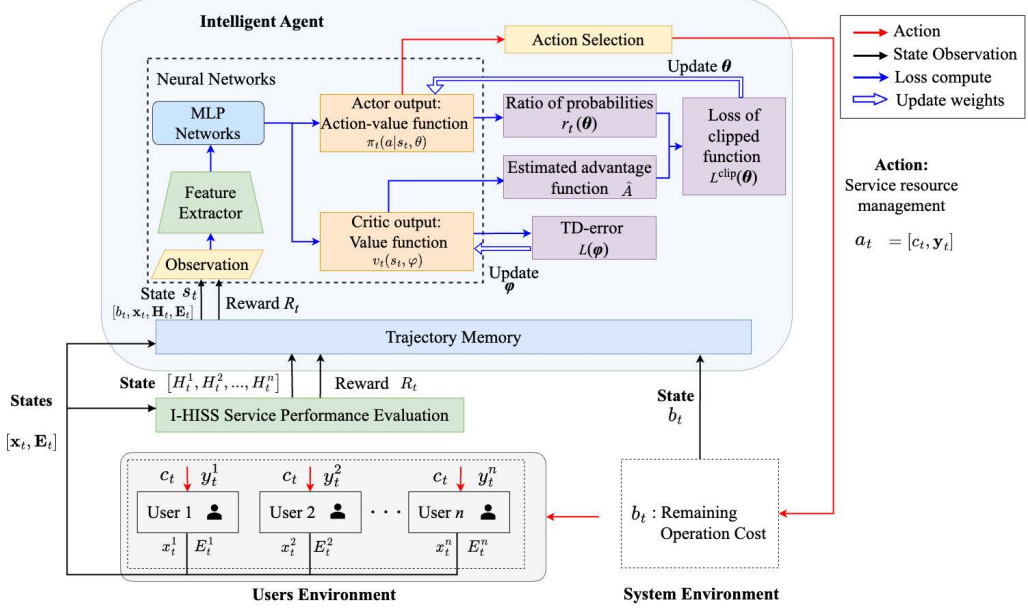

**Source(s):** Authors' own work

Figure II. The framework of the proposed algorithm for the service resource management problem with the agent and environment interaction (Arrows indicate the learning process of the algorithm).

*4.1 Agent and action space*

The I-HISS makes the service resource management decision to serve $n$ users. The action $\boldsymbol{a}_t = [c_t, \boldsymbol{y}_t] \in \mathcal{A}$, is a vector of actions consisting of the three components: service capacity $c_t$, service provision $\boldsymbol{y}_t$. The total size of the action space is $m \cdot (2^n + 1)$. To overcome the "curse of dimensionality" for the resource allocation problem, we use the deep reinforcement learning approach that represents the learnt functions as a neural network to tackle this problem.

*4.2 Environment and states*

The $n$ users constitute the environment in the deep reinforcement learning approach, which interacts with the learning agent and co-creates the health outcome. State $\boldsymbol{s}_t = [b_t, \boldsymbol{x}_t, \boldsymbol{H}_t, \boldsymbol{E}_t]$, where $\boldsymbol{s}_t \in \mathcal{S}$ is a vector at time $t$. Each component is characterized as follows:

- $b_t \in \mathbb{R}_+$: the balance of available operation costs at the decision point $t$, which depends on the action $\boldsymbol{a}_t$ and the previous state. Nonnegative balance $b_t \geq 0$ is considered as a constraint in the deep reinforcement learning approach. The available operation cost balance $b_{t+1}$ will change in accord with the service input over time $t$, where $b_{t+1} = b_t - c_t \geq 0$, $b_0 = B$. If the $b_{t+1} < 0$, the next actions $c_{t+1}$ will be equal to zero until the end of the service period.

- $\boldsymbol{x}_t \in \mathbb{R}_+^n$: $\boldsymbol{x}_t = (x_t^1, x_t^2, \ldots, x_t^n)$, where $n$ is the number of users, represents the engagement behaviour of each user based on the individual's inner behaviour motivation.

- $\boldsymbol{H}_t \in \mathbb{R}_+^n$: the health outcome of each user, which represents the value co-creation by the I-HISS and the user, where $\boldsymbol{H}_t = (H_t^1, H_t^2, \ldots, H_t^n)$, $n$ is the number of users. The transition of the health outcome $H_{t+1}^i$ for the user $i$ is calculated using Eq. (2).

- $E_t \in \xi \cdot \mathbb{R}_+^n$: additional information related to each user is collected by the heterogeneous IoT devices. The motivations for human engagement behaviour are intricate. While attempting to forecast the user's state, multiple factors need to be considered, such as the user's fatigue level and sleep state. We assume that $\xi$ types of indicators are collected in the I-HISS.

Hence, the state space is a $(\xi n + 2n + 2)$-dimensional vector. Deep reinforcement learning approach uses the neural network structures to deal with the discrete and quite large state space.

*4.3 Reward function*

The reward function is defined as the change in the health outcome value when action $a_t$ is taken at state $s_t$ and arrives at a new state $s_{t+1}$, as shown below.

$$R(s_t, a_t, s_{t+1}) = \sum_{i=1}^{n}(H_{t+1}^i - H_t^i) = \sum_{i=1}^{n} y_t^i (x_t^i)^{\alpha_1} \left(\frac{c_t}{\sum_{i=1}^{n} y_t^i}\right)^{\alpha_2}. \quad (4)$$

The I-HISS aims to find a decision rule for adaptive service resource management that maximizes the total health outcome for $n$ users. The function $\pi$ maps each state $s_t \in \mathcal{S}$ to an action $a_t \in \mathcal{A}$, is considered as a policy in this paper. Therefore, a policy $\pi$ is a distribution over actions given states:

$$\pi(a_t|s_t) \doteq \mathbb{P}[a_t|s_t], \forall a_t \in \mathcal{A}, s_t \in \mathcal{S}, t = 0,1,\ldots,T. \quad (5)$$

$v_\pi(s_t) = \mathbb{E}_\pi[R_t + v_\pi(s_{t+1})|s_t], \forall s_t \in \mathcal{S}$ is denoted as the value function for policy $\pi$, where $\mathbb{E}_\pi[\cdot]$ denotes an expected value, $v_\pi(s_{t+1})$ denotes the value function for the next state $s_{t+1}$. Solving the resource management problem necessitates the finding of optimal policies $\pi^*$ that yield the greatest return through service. There exists an optimal policy $\pi^*$ that is superior to or equal to all other policies, $\pi^* \geq \pi, \forall \pi$, where:

$$\pi^* \doteq \arg\max_\pi v_\pi(s_t), \quad \forall s_t \in \mathcal{S}. \quad (6)$$

*4.4 Algorithm description*

As depicted in **Source(s):** Authors' own work

Figure *II*, the deep reinforcement learning algorithm consists of two modules: the "actor" and "critic" networks, which control action selection and evaluate the actor's actions, respectively. The actor network with weights $\theta$ learns the parameterized policy $\pi(a_t|s_t; \theta)$ while the critic network with weights $\varphi$ evaluate the current parameterized policy $\pi(a_t|s_t; \theta)$ by computing the state-value function $v(s_t; \varphi)$. Additionally, during the learning process, the PPO-clip imposes a clip interval on the probability ratio term in the objective function to eliminate incentives for the new policy to diverge from the old policy. In this subsection, we illustrate the structure of hybrid neural networks, as well as the learning process for the algorithm.

*4.4.1 Hybrid neural networks.* Figure III illustrate the structure of hybrid neural networks with shared and diverging network to overcome the restrictions of the high-dimensional state and action space in the paper. The shared structure includes of an input layer, a single long-short term memory (LSTM) layer, and a normalization layer. Each memory cell in the LSTM layer regulates the flow of information into and out of the memory cells, enabling selective retain of relevant information over extended periods (Jia *et al.*, 2022). The LSTM layer is used for predicting user behaviour, and layer normalization applied to normalize activation of the LSTM layer across different input dimensions. The diverging network has two layers of multi-layer perceptions (MLPs) and a dense layer with a rectified linear activation function (ReLU) that outputs the action function for the actor network. The value function is produced by the critic network, which comprises of three MLP layers and a dense layer with ReLU.

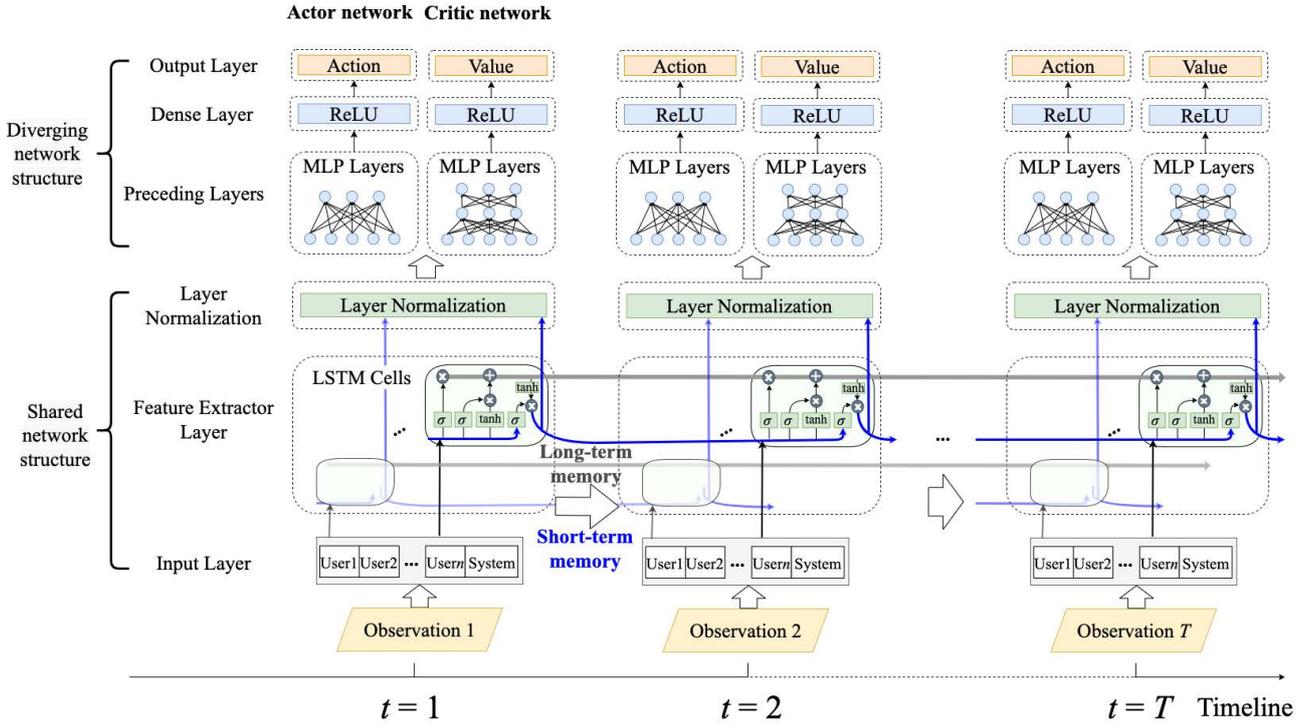

**Source(s):** Authors' own work

Figure III. The structure of the proposed hybrid neural networks with the long-term memory flow (grey arrows) and short-term memory flow (blue arrows)

*4.3.2 Learning process.* The critic network can be trained through gradient descent to minimize the mean square error of the temporal difference error $L(\boldsymbol{\varphi})$, calculated by $L(\boldsymbol{\varphi}) = \left(R_t - v(\boldsymbol{s}_t; \boldsymbol{\varphi})\right)^2$. The "Actor" then updates the policy distribution $\pi(a_t|\boldsymbol{s}_t; \boldsymbol{\theta})$ in the direction suggested by the "Critic" by maximizing an objective function $L^{\text{clip}}(\boldsymbol{\theta})$ with respect $\boldsymbol{\theta}$, where:

$$L^{clip}(\boldsymbol{\theta}) = \widehat{\mathbb{E}}_t\big[\min r_t(\boldsymbol{\theta})\,\hat{A}_t, clip(r_t(\boldsymbol{\theta}), 1-\epsilon, 1+\epsilon)\hat{A}_t\big], \qquad (7)$$

where $\widehat{\mathbb{E}}_t[\cdot]$ represents the empirical expectation over timesteps. The probability ratio $r_t(\boldsymbol{\theta})$ between old policy and the new policy is defined as $r_t(\boldsymbol{\theta}) = \pi(a_t|\boldsymbol{s}_t; \boldsymbol{\theta})/\pi(a_t|\boldsymbol{s}_t; \boldsymbol{\theta}_{\text{old}})$. The function $clip(r_t(\boldsymbol{\theta}), 1-\epsilon, 1+\epsilon)$ clips the ratio

$r_t(\theta)$ that falls inside the interval $[1 - \epsilon, 1 + \epsilon]$, where $\epsilon$ is a hyperparameter. $\hat{A}_t$ is the estimated advantage at time $t$, using the state-value function $v_\pi(s_t)$ as a baseline for evaluating an action for current policy $\pi$, is given as follows.

$$\hat{A}_t = q_\pi(s_t, a_t) - v_\pi(s_t). \tag{8}$$

The function $L^{\text{clip}}(\theta)$ minimizes the normal objective and clipped goal to avoid large deviation from the old policy, hence enabling the PPO algorithm to improve the training stability of the policy networks. The pseudo-code of the PPO-Clip algorithm can be found in Schulman *et al.*,(2017).

## 5. Experiments

This study aims to understand how optimal service resource management strategy can be applied in the IoT-based health information service to improve users' service satisfaction. A simulated example was run to test and fine-tune the proposed algorithm, as well as verify the effectiveness, sensitivity of the framework in health information service. After that, the proposed optimal service resource management strategy was implemented in business.

*5.1 A practical example (simulated) with different users' behaviour assumptions*

We ran and visualized simulations in Python using TensorFlow on a personal computer with the Intel (R) Core (TM) i7 CPU with 32.00 GB of RAM. With the simulation, we have the lifelog data of 16 participants' daily activities obtained between January 11, 2019 and March 30, 2020. There was a total of 16 participants, twelve men and three women, between 25 to 60 years old, with an average age of 34 years. The diversity of training and exercise background among the participants improved the validity of the dataset. Some are active athletes, some previous athletes, and some rarely exercise at all. The dataset was obtained from multiple sources, including Fitbit Versa 2 smartwatch wristbands, the PMSys sports logging smartphone application. The Fitbit Versa 2 fitness smartwatch was employed to collect objective biometrics and activity data. Participants were encouraged to consistently wear the watch. During training sessions, they logged in using the exercise menu option to select a specific activity, such as running or using the treadmill. For subjective assessments, the PMSys sports logging smartphone application was utilized to record participants' wellness, training load. Wellness reports were submitted through a series of questionnaires, and training load (calculated as Session Rating of Perceived Exertion - sRPE) was reported after each training session. Additional details regarding the data source can be found in Thambawita *et al.*, (2020). We evaluated the data quality based on the data consistency, validity, and accuracy. The consistency of the dataset was maintained by covering the reporting period from November 2019 to March 2020. The diversity of training and exercise background among the participants improved the validity of the dataset. In terms of data accuracy, the sports-activity data was automatically recorded by Fitbit Versa 2 smartwatch wristbands. A limitation of this dataset is the presence of noise resulting from its time-series data, which requires addressing missing data and identifying outliers. The proposed method in this study involves using a neural network to generate estimations for missing data. Another constraint of the dataset involves addressing technical obstacles related to the integration of heterogeneous data sources. We selectively utilised specific indicators relevant to our research assumptions rather than using the whole dataset. The assessment user engagement behaviour in our study was based on the subjective readiness levels in wellness reported by the participants.

Table I provides the units of the corresponding metrics to describe the indicators we selected in real world situation in the health information service. The performance of the value co-creation is calculated according to the engagement behaviour of the participant in health management. The indicator named "readiness" in the PMSys can be considered the degree of the participant's engagement motivation. The "readiness" has a 0-10 scale representing how ready the participant is to engage in the healthcare activity, where a score of 0 means not ready at all, and a score of 10 indicates that they are completely up for the healthcare activity. Except for the "readiness", we select four kinds of indicators related to the individual's daily exercise by combining the sports-activity data and the lifelogging data from the heterogeneous IoT devices, including the "calories" collected by the Fitbit Versa 2 smartwatch, the "fatigue" and "mood" extracted from the PMSys, and sRPE calculated from the product of the training load and the reported rating of perceived exertion. We performed a correlation analysis to confirm that these indicators correlated with "readiness" (i.e., effort level). The correlations between readiness level and the other indicators can be calculated using Pearson correlation. The results showed a significant correlation between readiness level and other indicators, particularly the "fatigue" level and the "mood" level.

Table I. The units of the corresponding metrics in the experiments and data statistics

| Measure | Indicator | Description | N | Min | Max | Mean | Std. |
|---|---|---|---|---|---|---|---|
| **Consumer engagement behaviour** | Readiness | Upload by the user, degree of consumers engagement motivation, on a scale of 1 to 10, convert to value (from 1 to 10) | 1658 | 1 | 10 | 5.17 | 1.66 |
| **Additional information related to consumers behaviour** | Calories | Calculate by Fitbit, total calories burned for the day | 2170 | 1374 | 11185 | 3044 | 867 |
| | Fatigue | Upload by the user, degrees of perceived fatigue, on a scale of 1 to 5, convert to value (from 1 to 5) | 1729 | 1 | 5 | 2.71 | 0.62 |
| | Mood | Upload by the user, degrees of mood, on a scale of 1 to 5, convert to value (from 1 to 5) | 1728 | 1 | 5 | 3.20 | 0.63 |
| | sPRE | Training load of perceived exertion, a metric calculated from the product of the session length and the reported rating of perceived exertion | 783 | 30 | 1800 | 379 | 265 |

**Source(s):** Authors' own creation/work

*5.1.1 Scenario of the experiments.* The experiments simulated user reactions to the health information service, consistent with prior empirical studies exploring the effects of using health information service to enhance user engagement levels. Our study considered four real-world user behaviour changes with the health information service.

- *Emulator 1 (E1)* - No significant effects were observed for health information service in increasing user behaviour (Kirwan *et al.*, 2012; Sirriyeh *et al.*, 2010). We assume that user engagement levels would remain stable during the service.
- *Emulator 2 (E2)* - A slightly increase in user engagement levels in health information service was observed (Voth *et al.*, 2016). We assume that user engagement levels would rise by 40% after one month.
- *Emulator 3 (E3)* - A highly increase in user engagement levels in health information service was observed (Muntaner *et al.*, 2016). We assume that there would a two-fold increase after one month in users' engagement levels.
- *Emulator 4 (E4)* - A decline and an increase in user engagement levels in health information service was observed (Gonze *et al.*, 2020). We assume that users' engagement levels would decline by 20% for one month and then increase by 60% after two months.

*5.1.2 Performance analysis.* The data were divided into 80% for training and 20% for testing. We converted 16 alternative service capacity scales to values ranging from 0 to 15. We conducted a 30-day simulation to explore hypothetical changes in user behaviour scenarios (Emulator E1, E2, E3, and E4). After tuning the hyperparameters in neural networks and reinforcement learning, we trained the proposed service resource management strategy. We measured computational complexity and average speed to evaluate the proposed algorithm. Table II shows that our proposed algorithm can be executed in 3 seconds during testing, which is close to real-time.

Table II. Training time and testing time of proposed algorithm and the competitor algorithms

| Algorithm | Training stage | | Testing stage |
|---|---|---|---|
| | Running time (seconds) | Epochs | Inference time (second) |
| PPO-Clip-Hybrid (Proposed Approach) | 2200 | 45000 | 2.6 |
| A3C-Hybrid | 2500 | 46000 | 2.8 |
| TRPO-Hybrid | 2900 | 80000 | 2.7 |
| PPO-Clip-MLP | 2800 | 200000 | 3.1 |
| A3C-MLP | 2600 | 450000 | 2.9 |
| TRPO-MLP | 2800 | 500000 | 2.3 |

**Source(s):** Authors' own creation/work

We compared our algorithm with a wide variety of algorithms. The competitor algorithms are implemented on the OpenAI-Baseline. Each simulation was run 100 times for statistical purposes. The competitor algorithms with two different models using the same hybrid neural network, which are synchronous Advantage Actor-Critic (A3C) and Trust Region policy Optimization (TRPO) (Plaat, 2022). On the other hand, instead of using hybrid neural network, we tested these algorithms using the multi-layer perceptions (MLPs) neural network for training. Table II shows that our proposed algorithm consistently outperforms in convergent speed compared with the algorithms using the MLP neural network structure. This indicates that the hybrid neural network architecture is useful for learning the users' engagement behaviour in the I-HISS. It's worth noting that the A3C-hybrid algorithm performs about the same as our proposed algorithm. Testing the experiments many times, we found our proposed algorithm is more stable than the A3C-Hybrid algorithm. The proposed deep reinforcement learning algorithm is a promising solution for service resource management in healt[...]

the rec[...]

*5.2 Fin[...]*

*5.2.[...]*

kinds o[...]

(2) deli[...]

equal distribution and delivering the healthcare information service to every participant in every time (fixed plan 3).

In order to improve the understanding of different service resource management plans, we take the "FitTime" compan[...]
particip[...]
consulta[...]
allocati[...]
availabl[...]
particip[...]
where t[...]
plan 3 allocates service capacity equally among all participants. Each user is allowed to get a total of 10 minutes of consultation time for each consultation service in this plan.

**Source(s):** Authors' own work

Figure *IV* shows that our proposed algorithm outperformed fix strategies. This is because the service provisioning strategy providing health information service to inactive participants cannot produce value in the health information service system. Service adaptation strategy reduces operation costs and support sustainable operation of the I-HISS by dynamic control service capacity. Moreover, A comparison of the results for E1, E2, E3, and E4, indicates that the performance in the I-HISS has no significant vary across different scenarios. Therefore, our algorithm can adapt to the users' engagement behaviour to some extent. Compared to other service resource management strategies, our proposed algorithm improved service performance regardless of user behaviour. In fixed plan 1, the high number of online users during peak hours results in lower quality of consultation services for each user. In fixed plan 2, although each user can access services, the low number of online users during off-peak hours leads to coach idleness. Fixed plan 3 maintains

consistency in service capacity and delivery but fails to meet the dynamic health needs of users. Therefore, our proposed algorithm dynamically adapts to users' changing health demands, allocating more service resources to active users during peak periods to enhance their service experience. It reduces coach scheduling during off-peak hours, thereby lowering operational costs for the system. The co-creation of service performance between the service provider and the users influences uses' service experience and ultimately expands the marketing share of the IHISS will expand accordingly to the quality of their users' service experiences.

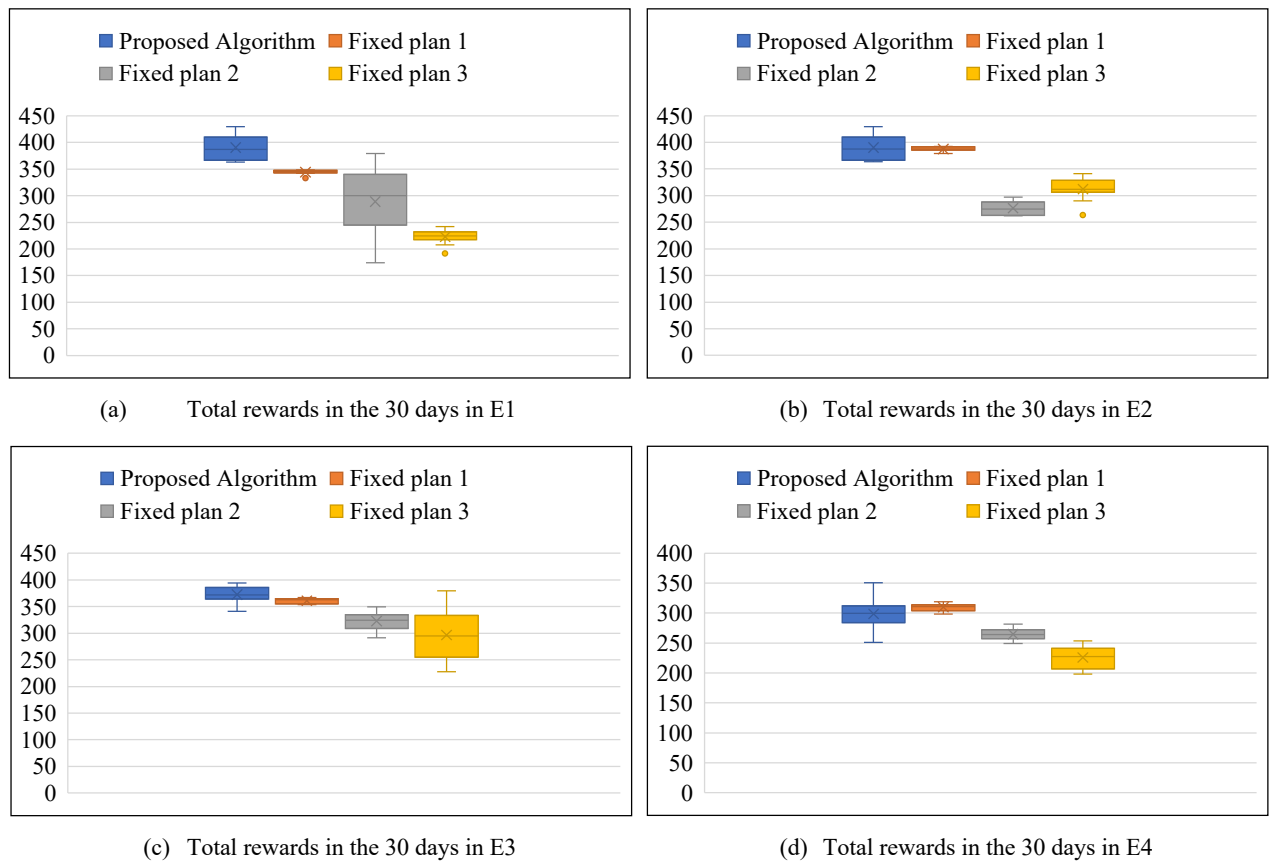

Source(s): Authors' own work

Figure IV. Performance of proposed algorithm comparing with fixed service resource management strategies ($\alpha_1 = 0.5, \alpha_2 = 0.4, B = 100, \beta = 0.9$)

*5.2.2 Improvement in the personal benefits for the users.* The health information service has a positive impact on the improvement of individual benefits for users. Users exert effort and engage in behaviours to complete health-related tasks, which can improve their health. Table III demonstrates the personal advantages for users under the adaptive service resource management strategy compared with other service resource management strategies. The same as the former discussion, other service resource management strategies are service capacity equal distribution in every time (fixed plan 1), delivering the healthcare information service to every participant in each time (fixed plan 2), and service capacity equal distribution and delivering the healthcare information service to every participant in every time (fixed plan 3). Sixteen users were chosen for the test. When a service provider adopts the adaptive service resource management strategy, the health outcomes of all users increase by more than 7.11 %. The active participation of users and service provider in the service process contributes to the creation of value for both parties.

Table III. Average personal total rewards with the proposed service resource management strategy and other service resource management strategies for 30-day simulation

| NO. | IHISS | Fixed Plan 1 | Fixed Plan 2 | Fixed Plan 3 | NO. | IHISS | Fixed Plan 1 | Fixed Plan 2 | Fixed Plan 3 |
|---|---|---|---|---|---|---|---|---|---|
| **P1** | 42.3 | 39.1 8.18% | 21.3 98.59% | 20.6 105.34% | **P9** | 14.8 | 12.9 14.73% | 8.1 82.72% | 10.3 43.69% |
| **P2** | 21.5 | 19.5 10.26% | 17.2 25.00% | 15.2 41.45% | **P10** | 25.4 | 23.4 8.55% | 21.3 19.25% | 20.7 22.71% |
| **P3** | 32.1 | 29.4 9.18% | 28.3 13.43% | 28.7 11.85% | **P11** | 30.2 | 28.8 4.86% | 25.6 17.97% | 23.9 26.36% |
| **P4** | 13.8 | 11.2 23.21% | 9.1 51.65% | 10.5 31.43% | **P12** | 16.5 | 14.6 13.01% | 9.2 79.35% | 12.8 28.91% |
| **P5** | 19.4 | 17.3 12.14% | 13.2 46.97% | 11.4 70.18% | **P13** | 26.2 | 23.5 11.49% | 21.5 21.86% | 20.2 29.70% |
| **P6** | 28.7 | 25.8 11.24% | 24.1 19.09% | 26.1 9.96% | **P14** | 19.2 | 17.9 7.26% | 12.1 58.68% | 14.5 32.41% |
| **P7** | 12 | 10.3 16.50% | 5.3 126.42% | 7.2 66.67% | **P15** | 27.1 | 26.4 2.65% | 23.8 13.87% | 21.9 23.74% |
| **P8** | 22.6 | 21.1 7.11% | 18.4 22.83% | 17.2 31.40% | **P16** | 11.1 | 9.9 12.12% | 9.4 18.09% | 9.5 16.67% |

(a) Total rewards for different $\alpha_1$

(b) Total rewards for different $\alpha_2$

(c) Total rewards for different $B$

(d) Total rewards for different $\beta$

**Source(s):** Authors' own creation/work

The former two subfigures in

**Source(s):** Authors' own work

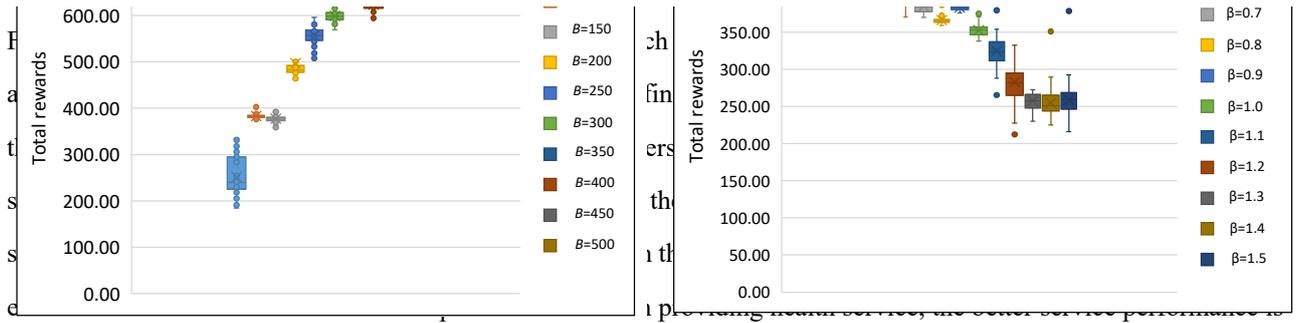

(a) Total rewards for different $\alpha_1$      (b) Total rewards for different $\alpha_2$

(c) Total rewards for different $B$      (d) Total rewards for different $\beta$

**Source(s):** Authors' own work

Figure V show that increasing the service operation budget $B$, and decreasing the cost elasticity of the service capacity input $\beta$ improves the average performance of the I-HISS. However, it is noteworthy that the value co-creation outcome dose not increase proportionally with the service capacity input. This may be due limitations imposed by user engagement behaviour.

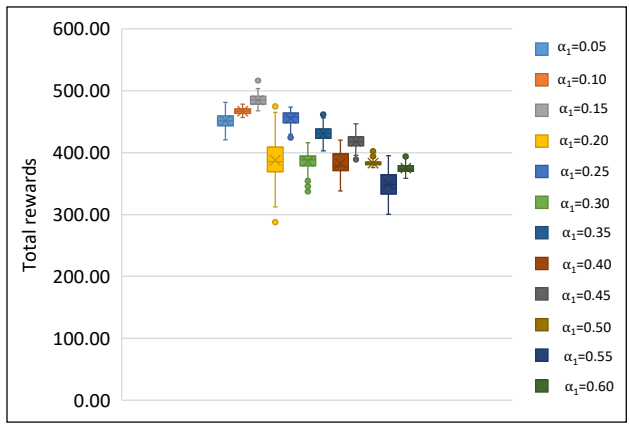

(a) Total rewards for different $\alpha_1$

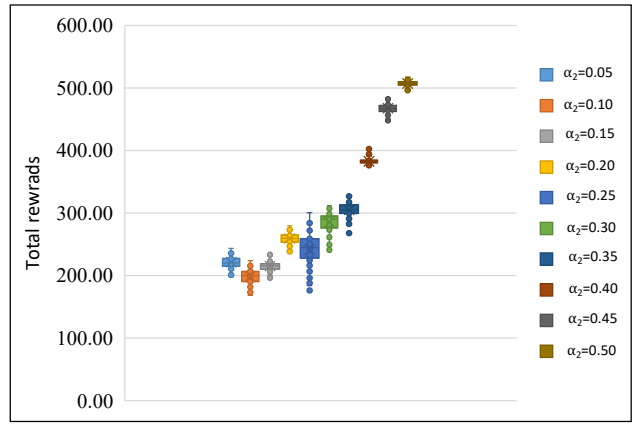

(b) Total rewrads for different $\alpha_2$

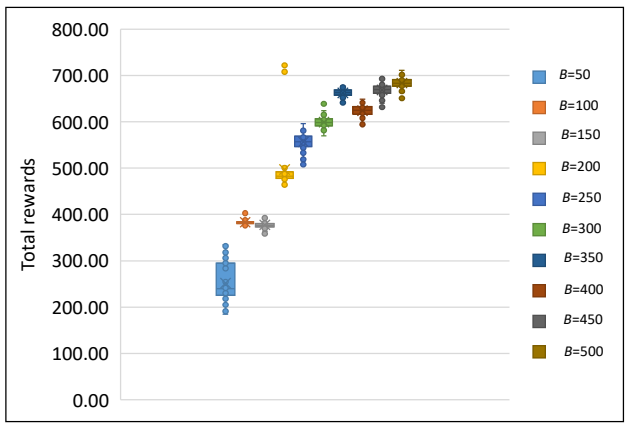

(c) Total rewards for different $B$

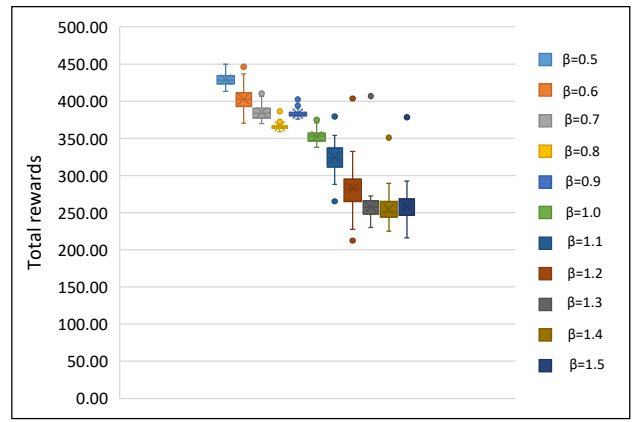

(d) Total rewards for different $\beta$

**Source(s):** Authors' own work

Figure V. Total rewards for different parameters testing in Environment E2 ($\alpha_1 = 0.5, \alpha_2 = 0.4, B = 100, \beta = 0.9$)

Source(s): Authors' own workFigure V indicates that although users are crucial to value co-creation, their influence on service outcome may be limited due to the fact that the I-HISS can adjust its operations based on users' engagement behaviour. The performance of the health information system improves as the I-HISS assumes a greater role in the process of value co-creation. The average performance of I-HISS is enhanced over time by increasing the service budget or decreasing the cost elasticity of service capacity. As users become more involved in the service process, their influence on the service outcome increases. Depending on the users' level of expertise and their perception of the value of co-creation, this increase may have positive or negative effects. Users collaborate with the service provider to develop the health information service that meets users' fitness needs, and after using the service, users can provide feedback and suggestions to improve the service.

## 6. Conclusion and implications

This study explores a novel IoT-based health information service system(I-HISS) that has been seldom considered in previous research. We analysed service management in I-HISS by service resource allocation to its subscribed users. At first, we considered the preferences of user behaviour and different service demands. We analyse the fitness data of users, predicting their engagement behaviour, and quantifying the performance of I-HISS's service. Second, in order to guarantee the affordable optimal user service experience, we developed a deep reinforcement learning algorithm to dynamically allocate service resource to adapt to fluctuations in user demands. In addition, we provided managerial perspective to improve the strategies for service resource allocation based on the dominant factor level of the user engagement in generating service value in different business situations. The conclusions and managerial insights can be summarized as follows.

(1) It is important for I-HISS to make precise predictions about user exercise preferences by collecting health data using smart hardware. This decreases the impact of user behavioural uncertainty on service outcomes, maintaining the stability of service performance.

(2) An adaptive approach is the most effective strategy for service resource management for I-HISS. I-HISS adopts a dynamic approach to flexibly control and allocate service quality, resulting in significant improvements in service performance and effectively addressing the fluctuating of users' health demands during both high-demand and low-demand periods. This strengthens the user service experience and reinforces the company's competitiveness in the market.

(3) The importance of user's fitness effort in health results has a relatively minor impact on service performance in I-HISS. For users with different levels of health guidance demands, I-HISS can optimize the service performance through the allocation of service resources.

(4) The significance of I-HISS health guidance in health improvement is greatly influential on service performances,

meaning that fitness results depend heavily on the guidance provided by the information service. Greater importance placed on I-HISS health guidance leads to improved performance in health services. I-HISS should emphasize service quality management in order to improve the user service experience.

(5) Increasing service budgets or decreasing unit service costs can improve I-HISS's service performance. However, the improvement in service performance is not proportional to the increase in service budgets. This could be attributed to the limited user engagement behaviour resulting in a waste of service resource.

This study presents an IoT-based health information service system (I-HISS) that delivers interactive health information service using heterogeneous IoT structures and a value co-creation service model to improve service performance. The I-HISS utilizes the IoT architecture to assist the information service and automatically allocate service resource based on the users' engagement behaviour. Our service resource management strategy with service-dominant logic of value co-creation process was validated by simulation experiments with real-world data to examine various user interaction trends. The findings show that the proposed deep reinforcement learning algorithm outperforms other algorithms in terms of problem-solving time and convergence steps, making it an efficient and effective solution for service resource management in health information service. The results also show that our service resource management strategy increases both service providers' business revenue and users' personal benefits in I-HISS. The users become more involved in the service delivery process, their influence on the outcome of the I-HISS increases. However, this increased control may have positive and negative effects, depending on users' level of expertise and perceived value of co-creation. The results have significant managerial implications for service providers to improve their health information service and user engagement.

There are certain limitations in our study. First, we only consider the interaction between I-HISS and a single user. However, user may connect with each other by sharing their exercise progress, accomplishments, and encouraging one another. These interactions can have beneficial consequences on user engagement and motivation, resulting in improved health results. Second, we ignore the degree of health knowledge attained by users, assuming that the effect of health service on user's health improvement is equivalent. In subsequent research, we plan to extend this study by including diverse user groups with different level of learning ability in fitness service. In addition, our study will combine social interactions among users such as likes and shares. Thus, potential future research will use the graph neural networks technology for analysing user social support in exercise, with the objective of maximizing the improvement of users' health conditions. Third, we can use behavioural experiments to test the model parameters of health service for user health improvement, making the theoretical findings of this study more practical for health industry.